\pdfoutput=1

\documentclass[11pt]{article}

\usepackage[]{ACL2023}

\usepackage{times}
\usepackage{latexsym}

\usepackage[T1]{fontenc}

\usepackage[utf8]{inputenc}

\usepackage{microtype}

\usepackage{inconsolata}

\title{Foundation Models for Low-Resource Language Education (Vision Paper)}

\author{Zhaojun Ding$^*$, Zhengliang Liu$^*$, Hanqi Jiang$^*$ \\
{\bf Yizhu Gao}, {\bf Xiaoming Zhai}, {\bf Tianming Liu} \\
{\bf Ninghao Liu} \\
University of Georgia \\
\texttt{\{zhaojun.ding, zl18864, hj67104, yizhu.gao, xiaoming.zhai, tliu, ninghao.liu\}@uga.edu}\\[2pt]
\textsuperscript{$^*$}These authors contributed equally to this work.}

\begin{document}
\maketitle
\begin{abstract}
Recent studies show that large language models (LLMs) are powerful tools for working with natural language, bringing advances in many areas of computational linguistics. However, these models face challenges when applied to low-resource languages due to limited training data and difficulty in understanding cultural nuances. Research is now focusing on multilingual models to improve LLM performance for these languages. Education in these languages also struggles with a lack of resources and qualified teachers, particularly in underdeveloped regions. Here, LLMs can be transformative, supporting innovative methods like community-driven learning and digital platforms. This paper discusses how LLMs could enhance education for low-resource languages, emphasizing practical applications and benefits.
\end{abstract}

\section{Introduction}

Recent research in large language models (LLMs) has demonstrated their profound capabilities in processing and generating natural language, leading to advancements across various domains of computational linguistics~\cite{achiam2023gpt,ouyang2022training,touvron2023llama,team2024gemma,jiang2023mistral,bai2023qwen}. 
Despite these advancements, applying LLMs in the study of low-resource languages presents significant challenges. These models often struggle with the limited availability of training data, which can lead to poorer performance and less effective language understanding and generation. Additionally, the complexities of accurately capturing the nuances and cultural contexts of less-documented languages further complicate their use in this area.
To tackle the problem, recently multilingual models have been studied, extending language models to low-resource languages~\cite{chang2023multilinguality,qin2024multilingual}.

Education in the domain of low-resource languages often face significant challenges due to limited access to teaching materials, qualified instructors, and formal language programs~\cite{kohnke2023chatgpt}. Typically, these languages are spoken in regions where educational infrastructure is less developed, making traditional classroom-based learning difficult to implement. As a result, generative AI tools such as LLMs have the potential to revolutionize educational initiatives~\cite{lee2023multimodality,tian2024assessing,shu2024llms,latif2024systematic} such as community-led classes, oral tradition, and providing digital tools and automated platforms~\cite{latif2024systematic,jiang2024oraclesage,shu2024transcending,zhong2024evaluation,liu2023context,liu2024understanding,liu2023transformation,zhou2024comprehensive}. 
In this vision paper, we discuss various aspects in low-resource language education that could be enhanced by the recent foundation models.

\section{Foundation Model Development}
This section explores the development of multilingual foundation models designed to extend the capabilities of NLP systems to low-resource languages. Both large language models and vision-language models (VLMs) are discussed in the context of pre-training, fine-tuning, and in-context learning. These approaches aim to overcome the data scarcity challenges inherent in low-resource languages, while also supporting educational and practical applications.

\subsection{FM Pre-training}
Pre-training forms the foundational stage for both LLMs and VLMs, where models learn to extract general linguistic or multimodal patterns from large-scale datasets. LLM pre-training typically employs loss functions such as masked language modeling (e.g., BERT~\cite{devlin2018bert}) or autoregressive objectives (e.g., the GPT series~\cite{radford2018improving,radford2019language,brown2020language} and the Llama family of models~\cite{touvron2023llama,touvron2023llama2,dubey2024llama}). These techniques allow LLMs to capture complex syntactic, semantic, and contextual relationships in text. 

LLM pre-training commonly~\cite{liu2024understanding,zhou2024comprehensive,liu2023context} employs objectives such as:

\begin{itemize}
    \item \textbf{Masked Language Modeling (MLM):} This method masks random tokens in the input sequence and trains the model to predict these masked tokens during pre-training~\cite{devlin2018bert}. BERT and many of its variants use MLM as their primary pre-training objective, enabling bidirectional context understanding.
    
    \item \textbf{Autoregressive Modeling:} This approach trains models to predict the next token in a sequence by learning patterns from vast amounts of text data~\cite{brown2020language}. The model conditions each new token on all previously generated tokens, enabling it to maintain coherence and capture long-range dependencies while generating text.

     \item \textbf{Denoising AutoEncoder (DAE):} This technique corrupts the input data with various noise functions (masking, deletion, permutation) and trains the model to reconstruct the original sequence~\cite{lewis2019bart}. BART and T5 employ DAE-based objectives to learn robust text representations and generation capabilities.

     \item \textbf{Replaced Token Detection (RTD):} This task trains a discriminator to distinguish between original and synthetically replaced tokens in the input sequence~\cite{clark2020electra}. ELECTRA demonstrated that RTD enables more compute-efficient pre-training compared to MLM, particularly for smaller models.

     \item \textbf{Next Sentence Prediction (NSP) and Sentence Order Prediction (SOP):} NSP trains models to determine if two sentences appear consecutively in the original text~\cite{devlin2018bert}, while SOP focuses on predicting the correct ordering of consecutive sentences~\cite{lan2019albert}. ALBERT showed that SOP could be potentially more effective than NSP for learning inter-sentence coherence.
\end{itemize}

However, the representation of low-resource languages in existing datasets remains limited, often leading to suboptimal performance in these languages~\cite{magueresse2020low}.

For VLMs, pre-training involves aligning image and text embeddings within a unified semantic space. Techniques include the use of CLIP’s contrastive learning framework~\cite{radford2021learning}, where paired image-text data are used to maximize alignment, and latent diffusion models like Stable Diffusion~\cite{rombach2022high}, which focus on text-to-image tasks. Models such as LLaVA~\cite{liu2024visual} and Kosmos-2~\cite{peng2023kosmos} enhance cross-modal understanding, enabling tasks such as visual question answering and captioning. However, these models predominantly rely on high-resource language datasets, leaving low-resource languages underserved.

Targeted pre-training strategies can significantly benefit low-resource languages. Continuous pre-training on multilingual corpora or domain-specific datasets enhances model adaptability~\cite{xie2023efficient,liu2023summary}. For VLMs, incorporating culturally relevant visual and textual data ensures better alignment for underrepresented languages. In educational settings, such pretrained models can generate learning materials, including multimodal content, that cater to diverse linguistic contexts, ultimately improving accessibility and inclusion.

\subsection{FM Fine-tuning}
Fine-tuning adapts pre-trained models to specific tasks or domains by training them on task-specific data and optimizing loss functions to refine their performance. This process allows foundation models to address particular requirements, such as low-resource language processing or multimodal applications.

Fine-tuning typically uses a supervised learning approach, where the model’s predictions are compared against labeled data. The most commonly used loss function is cross-entropy~\cite{li2024entropic}, which measures the difference between predicted probabilities and the true labels. By minimizing cross-entropy loss, the model learns to generate accurate outputs for the target task while mitigating overfitting through techniques like dropout, weight decay, or early stopping. Optimization algorithms, such as AdamW~\cite{loshchilov2017decoupled}, are often employed to fine-tune model parameters efficiently.

\paragraph{Instruction Fine-tuning for LLMs and VLMs:} Instruction fine-tuning adapts pre-trained models to align with specific instructions and tasks, enhancing their generalization to unseen scenarios. For LLMs, this involves training on datasets containing task-specific instructions paired with responses. This approach, as seen in models like InstructGPT~\cite{ouyang2022training} and FLAN-T5~\cite{chung2024scaling}, enables effective zero-shot and few-shot task generalization. Such fine-tuning often leverages datasets encompassing diverse tasks, such as answering questions, performing translations, and summarizing text. For low-resource languages, instruction-tuning can be pivotal. By training on carefully curated datasets of low-resource languages (e.g., Quechua or Cherokee), LLMs can address linguistic gaps and support education through task-based language exercises tailored to underrepresented communities.

In the context of VLMs, instruction fine-tuning expands the capability of models to follow multimodal instructions. Recent developments like LLaVA~\cite{liu2024visual} and Flamingo~\cite{alayrac2022flamingo} have demonstrated how visual instruction-tuning aligns language and vision encoders with task-specific goals, such as image captioning and visual question answering. For instance, by fine-tuning on datasets with paired image-text instructions, these models can describe images in underrepresented languages or perform tasks such as cultural-specific object identification. This is particularly impactful for education, where VLMs can enable immersive learning experiences, providing visual aids and engaging storytelling in local languages.

\paragraph{Applications in Education:}
Fine-tuned models open up significant possibilities for personalized and localized educational tools. Instruction-tuned LLMs can assist students in learning low-resource languages by generating culturally specific exercises, correcting grammar, or providing contextually appropriate examples. Similarly, fine-tuned VLMs can enable immersive learning experiences by creating visual aids, interactive storytelling, or image-based quizzes tailored to the learner’s linguistic and cultural background. These adaptations bridge the gap between global AI technologies and local educational needs, fostering inclusivity and accessibility in resource-constrained environments.

\subsubsection{Preference Alignment with RLHF}
Preference alignment refines foundation models by aligning their outputs with human preferences, enhancing their utility for specific applications. Two prominent approaches for preference alignment are Proximal Policy Optimization (PPO)~\cite{schulman2017proximal} and Direct Preference Optimization (DPO)~\cite{rafailov2024direct}, each with unique strengths.

\paragraph{Proximal Policy Optimization (PPO):}
PPO is a reinforcement learning method central to Reinforcement Learning from Human Feedback (RLHF)~\cite{kaufmann2023survey,ouyang2022training}. This process involves training a reward model on human-labeled rankings of model outputs, followed by fine-tuning the foundation model to maximize these rewards. PPO ensures stable updates to the model while aligning it with desired stylistic and behavioral traits, such as cultural sensitivity or conversational norms.

For low-resource languages, PPO can adapt models to handle specific cultural contexts or linguistic subtleties, making it suitable for applications like culturally aware chatbots or educational AI companions. However, the approach requires substantial human-labeled data for training the reward model, which may pose challenges in low-resource settings.

\paragraph{Direct Preference Optimization (DPO):}
DPO provides a more streamlined alternative by bypassing the need for a reward model and reinforcement learning loop. Instead, it directly fine-tunes the model on preference-labeled data using a contrastive loss to maximize the likelihood of preferred outputs over less-preferred ones. This simplicity reduces computational overhead and eliminates the instability risks inherent in PPO.

A significant advantage of DPO is its ability to reduce labeling costs. Unlike PPO, which requires ranked outputs for reward model training, DPO leverages binary or limited comparative preference data, minimizing the burden of annotation. This efficiency makes it particularly valuable for low-resource languages, where obtaining extensive labeled datasets is often infeasible. For instance, DPO can be used to train models that generate culturally sensitive text or conversational agents with minimal labeled examples.

\paragraph{Applications in Low-Resource Language Education:}
Both PPO and DPO can enhance educational tools for low-resource languages, but DPO's efficiency and lower data requirements make it especially impactful. By reducing the need for extensive labeling, DPO enables the rapid adaptation of AI systems to underserved languages and cultural contexts. In education, it supports the development of AI companions that provide personalized feedback, culturally sensitive guidance, or tailored lesson plans, fostering inclusivity and accessibility in resource-constrained environments.

In low-resource languages, RLHF can adjust models to reflect cultural sensitivity, conversational norms, and localized communication styles. Applications include AI learning companions tailored to specific linguistic contexts, culturally aware educational chatbots, and training systems for language-specific customer service.

\subsection{FM In-context Learning}
In-context learning (ICL)~\cite{dong2022survey} allows pre-trained models to generalize to new tasks without additional parameter updates by leveraging few-shot examples provided as input prompts. This capability is particularly valuable for low-resource languages, where annotated data is scarce. Properly designed prompts enable LLMs to recognize patterns and generate relevant outputs, reducing the reliance on large labeled datasets.

For VLMs, ICL extends to multimodal prompts that combine images with textual instructions. This facilitates zero-shot or few-shot learning for tasks like image captioning or visual question answering in low-resource languages. Models like Kosmos-2~\cite{peng2023kosmos} and DeepSeek-VL~\cite{lu2024deepseek} demonstrate strong zero-shot capabilities, making them suitable for applications where task-specific data are unavailable.

The intuitive and flexible nature of ICL is a significant advantage in educational contexts. LLMs can be prompted to create exercises, translate content, or summarize lessons in low-resource languages, while VLMs can answer visual questions or provide contextual understanding of imagery. This adaptability supports educators and learners in resource-constrained environments, democratizing access to AI-driven educational tools.

\section{Language Education Modules}
With foundational models for low-resource languages, we can generate comprehensive lessons and learning materials spanning various aspects~\cite{magueresse2020low}, including vocabulary and pronunciation, grammar, interactive exercises, cultural integration, and video generation. 
This section will discuss methodologies for leveraging foundational models for various language education modules, demonstrating how foundation model agents can effectively support education in low-resource languages. 

\subsection{Vocabulary and Pronunciation}
To build vocabulary resources, the foundation model agents can harness their ability to understand and generate contextualized text in low-resource language~\cite{guo2024teaching}. More specifically, these model agents can produce:

\begin{itemize}
    \item \textbf{Word Lists:} Starting with foundational word sets, models can extrapolate to generate synonyms, antonyms, and contextually related terms. For instance, a low-resource language with a small dictionary can be expanded by inferring semantically linked words using multilingual embeddings in large language models.
    \item \textbf{Contextual Examples:} Learners often benefit from seeing new vocabulary used in meaningful contexts. Foundation models can generate diverse, practical examples by embedding target words within sentences, dialogues, or short narratives. These examples provide learners with a deeper understanding of how words function in real-life scenarios~\cite{cahyawijaya2024llms}.
    \item \textbf{Thematic Categorization:} Vocabulary lists can be organized into thematic areas like family, nature, or technology. Foundation models trained on multilingual data can identify domain-specific terms and present them in learner-friendly ways, such as flashcards or interactive quizzes.
\end{itemize}

Pronunciation mastery is often a hurdle in language learning, especially for learners of low-resource languages with limited audio examples. Foundation model agents equipped for speech and multimodal processing provide innovative solutions~\cite{li2024echopulse,lajszczak2024base}:

\begin{itemize}
    \item \textbf{Phoneme-to-Audio Generation:} Text-to-speech (TTS) systems based on foundation models can generate audio samples for specific words or phrases. Foundation models or fine-tuned systems can provide native-like pronunciation examples, even when trained on small low-resource language datasets.
    \item \textbf{Phonetic Guides and Feedback:} By integrating speech recognition, models can evaluate learners’ pronunciation. For instance, a learner speaking a word can receive feedback on accuracy, intonation, and stress patterns. This fosters self-paced learning without needing a native speaker's constant availability.
    \item \textbf{Regional Variations:} For low-resource languages with significant dialectical differences, foundation models can be fine-tuned to generate region-specific pronunciations, offering learners exposure to the diversity within the language. This is particularly valuable for languages where phonetic, tonal, or lexical differences between regions are substantial.
\end{itemize}

The TF-IDF (Term Frequency-Inverse Document Frequency)~\cite{aizawa2003information} method can enhance vocabulary and pronunciation learning by identifying words that are both frequent and contextually important. For vocabulary, TF-IDF can prioritize words that are common in a learner’s target domain (e.g., travel or business), helping learners focus on terms with high relevance. In pronunciation practice, TF-IDF can highlight less common but important words or those with regional variations, guiding targeted practice. This method ensures that learners engage with vocabulary that is both practical and specific to their needs, optimizing their learning experience.

Foundation model agents can also be improved through tokenization, a critical component for processing and understanding language. Tokenization is the process of breaking down text into smaller units, or tokens, which are processed by language models. This is essential for understanding how different words and phrases are constructed in low-resource languages, especially when dealing with agglutinative or morphologically complex languages. By tokenizing text properly, foundation models can capture linguistic nuances more accurately and generate vocabulary lists that reflect the language’s true structure, rather than overgeneralizing based on sparse data.

\subsection{Grammar}
Grammar is fundamental to mastering any language, yet low-resource languages often lack comprehensive instructional materials. The foundation model agents can automate the creation of grammar explanations, exercises, contextualized examples, etc., all tailored to the specific needs of learners. By harnessing the power of these foundation model agents, we can provide learners of low-resource languages with personalized, accessible, and engaging grammar content.

\begin{itemize}
    \item \textbf{Grammar Rules Extraction and Explanation:} Foundation models can analyze multilingual datasets to extract and explain grammatical patterns for low-resource languages. By identifying syntactic structures such as word order, tense, and subject-verb agreement, foundation models can generate grammar guides tailored to the unique rules of a given low-resource language. These models can produce detailed explanations of fundamental grammatical concepts, from simple noun-adjective agreements to more complex structures, offering step-by-step guidance for learners.
    \item \textbf{Cross-Lingual Grammar Transfer:} Foundation models can also facilitate cross-lingual grammar transfer by comparing the grammatical structures of related languages. For example, a learner who speaks English may benefit from grammar lessons comparing English and a low-resource language, focusing on shared and divergent grammatical features. This comparison allows learners to leverage their existing linguistic knowledge, accelerating their learning process.
\end{itemize}

\subsection{Interactive Exercises}
The foundation model agents could offer various types of interactive exercises that engage learners in practicing the language in a controlled environment, providing immediate feedback and adjustments based on their responses~\cite{xiao2023evaluating}. The flexibility of these exercises ensures that learners can practice different aspects of language acquisition in a way that suits their individual needs. Below are several specific use cases that illustrate how foundation model agents can generate interactive exercises for low-resource language learning.
\begin{itemize}
    \item \textbf{Matching Exercises for Vocabulary Learning:} Learners are presented with a set of vocabulary words and their corresponding images or definitions. The learner must match each word with its correct meaning or image. The foundation models adapt the difficulty based on the learner’s progress, offering more challenging words or a wider variety of images as they advance.
    \item \textbf{Dialogue Simulation Exercises for Pronunciation Learning:} The foundation models can simulate real-life conversations, where learners interact with an AI-based character in a scenario (e.g., ordering food at a restaurant). The learner's speech input is evaluated in real-time, and the system provides feedback on pronunciation, fluency, and appropriateness of responses. This type of exercise helps learners practice speaking in a practical, contextual setting.
    \item \textbf{Error Correction Exercises for Grammar Learning:} Learners are presented with sentences that contain common grammatical errors (such as subject-verb agreement or tense inconsistencies), and they are tasked with identifying and correcting them. Immediate feedback is provided, explaining why a particular correction is necessary.
\end{itemize}

\subsection{Culture Integration}
Integrating cultural education into language learning is essential, particularly for low-resource languages. Language and culture are deeply intertwined, and foundation model agents—spanning text, voice, and vision—can generate materials that highlight how language operates within its cultural context, including folklore, history, art, and traditions~\cite{li2024culturellm}.

Foundation model agents can produce stories, songs, or dialogues that reflect the cultural heritage of the language’s speakers, exposing learners to folklore and idiomatic expressions. These materials not only teach language but also reveal cultural values and worldviews. For instance, learners can engage with traditional narratives that embody cultural meanings.

Moreover, foundation model agents can generate historical content that provides context on the language’s evolution and its speakers’ experiences. They can translate or summarize historical texts, highlight significant events or figures, and showcase scientific or artistic contributions, helping learners understand the language’s cultural and historical roots.

Visual content created by these foundation model agents can complement text and voice lessons, featuring landmarks, art, and indigenous knowledge. Interactive maps, art galleries, and multimedia on local practices, like traditional medicine or sustainable agriculture, allow learners to connect the language to the community’s way of life.

By integrating culture into low-resource language learning, foundation model agents offer a comprehensive approach that connects language contents with cultural identity, enriching the learners’ understanding and experience.

\subsection{Video Generation}
In the context of low-resource languages, one of the most significant barriers to education is the lack of access to qualified educators and learning buddies. Foundation model agents offer solutions to this problem, particularly through the generation of video content featuring virtual teachers and learning buddies~\cite{xu2024foundation}. These resources provide immersive, interactive learning experiences that cater to the needs of learners in underserved linguistic communities.

\paragraph{Virtual Teachers for Immersive Learning:}
Virtual teachers are AI agents that deliver lessons tailored to a student’s level and progress. Using a combination of text-based foundation models, voice synthesis, and computer vision technologies, these virtual teachers can simulate human-like interactions. For instance, voice models can generate speech in the target language while video models display facial expressions and gestures that align with cultural context, helping students not only learn the language but also understand its social and cultural nuances.

These AI-driven teachers enable scalable and personalized education in areas where human teachers are scarce. By adjusting lesson complexity based on the learner’s proficiency, virtual teachers can introduce new vocabulary and improve pronunciation over time. Additionally, they can be multilingual, supporting low-resource language learners from diverse language backgrounds.

The integration of non-verbal cues like body language and facial expressions in video lessons further enhances communication, helping students associate spoken words with actions. This immersive approach aids both comprehension and retention, making virtual teachers a powerful tool for language acquisition in under-resourced areas.

\paragraph{Learning Buddies for Enhanced Engagement:}
In addition to virtual teachers, learning buddies—AI agents designed to interact and engage with students—play a crucial role in fostering motivation and providing ongoing support. Learning buddies can offer encouragement, answer questions, and guide students through practice exercises. These AI agents can hold conversations, giving learners opportunities to practice speaking and listening in dynamic, real-life contexts~\cite{nyaaba2024generative}.

Learning buddies also provide personalized feedback, offering corrections when necessary and reinforcing new concepts. This adaptive approach builds student confidence and accelerates language mastery. Furthermore, learning buddies can be customized to reflect cultural nuances, making lessons not only linguistically accurate but also culturally relevant, thereby deepening students’ connection to the language.

\section{Adaptive Learning in Language Models and Education}
Adaptive learning in low-resource language education should primarily focus on creating personalized learning experiences based on individual student knowledge levels and progress. A foundation model-based system could analyze each learner's current understanding and mastery of language concepts to recommend customized learning paths, content, and objectives. This personalization would ensure that students receive appropriately challenging material and targeted practice opportunities that align with their learning goals and current capabilities.

Such a system could continuously evaluate student performance to adjust these recommendations dynamically. When students would demonstrate mastery of certain concepts, the system might advance them to more challenging material. Conversely, if students were to struggle with particular aspects, the system could provide additional practice and support in those areas. This dynamic adjustment would help maintain an optimal learning pace for each student while ensuring thorough understanding of fundamental concepts.

Beyond individual learning paths, adaptive learning would also encompass the technical evolution of language models themselves. These models should be able to update and improve their knowledge base through mechanisms like continuous pre-training and fine-tuning, particularly important for low-resource languages where traditional training data might be limited. This dual nature - personalized education and model evolution - could create an effective framework for low-resource language education that responds to both individual learner needs and evolving linguistic landscapes.

\subsection{Content Generation and Adaptation}
\paragraph{Model Evolution:}
Traditional adaptive learning systems face limitations with pre-authored content and fixed question banks~\cite{larusson2014learning}. In contrast, foundation models for low-resource languages should be able to continuously evolve their knowledge base through interaction with new linguistic data. This evolution would allow dynamic content generation that could reflect current language usage patterns and cultural contexts.

\paragraph{Educational Content:}
The adaptive capability should extend to the generation of varied, contextually appropriate learning materials that could be adjusted based on learner needs. This would overcome the traditional limitation of manual content updates and restricted practice variety~\cite{aleven2016help}. The system could generate exercises, examples, and explanations in real-time that would maintain cultural relevance and support progressive language acquisition.

\paragraph{Individualized Content:}
A foundation model-based system could construct customized learning paths by analyzing each student's progress milestones and learning objectives. When a student's goal would be professional communication, the content sequence might emphasize relevant vocabulary and formal language patterns. For basic conversational goals, the path could prioritize common phrases and daily expressions. Such goal-oriented content sequencing would help ensure that learning materials align with each student's specific language objectives.

\subsection{Feedback Mechanisms}
\paragraph{Interactive Guidance:}
Unlike traditional systems constrained to predefined responses~\cite{ma2014intelligent}, foundation models should be able to provide nuanced, contextual feedback through natural language understanding. This capability would lead to more sophisticated error analysis and targeted remediation strategies for language learners.

\paragraph{Cultural Sensitivity:}
The feedback system should be able to incorporate cultural context awareness. This feature would ensure that corrections and suggestions align with both linguistic accuracy and cultural appropriateness. This approach would address the limitation of traditional systems that cannot recognize or respond to culturally influenced language usage patterns.

\paragraph{Personalized Feedback:}
The feedback mechanism should be able to adapt to each student's position on their learning path and current learning objectives. For students working toward reading proficiency, feedback could emphasize character recognition and comprehension strategies. For those focused on speaking skills, the system might provide more pronunciation guidance. This alignment between feedback and learning goals would help students progress more effectively toward their target language proficiency levels.

\subsection{Student Modeling}
\paragraph{Comprehensive Assessment:}
Traditional modeling approaches often struggle with complex learning trajectories~\cite{corbett1994knowledge,chi2011evaluation}. Foundation models should be able to track multiple dimensions of language acquisition at once, from vocabulary and grammar to pronunciation and cultural understanding. This comprehensive tracking would create more detailed learner profiles.

\paragraph{Adaptive Pathways:}
These systems should be able to analyze patterns in learner interactions and performance to adjust learning paths for individual needs. This approach would address the traditional limitation of insufficient consideration of affective states~\cite{reis2018affective} and motivation in learning. The result would be a more personalized and engaging educational experience.

\subsection{Integration Considerations}

\paragraph{Technical Implementation:}
When integrating foundation models for low-resource language education, scalability could potentially be a primary concern. The system architecture must support efficient model deployment while maintaining quick response times for interactive learning sessions. Additionally, quality control mechanisms must be implemented to ensure the accuracy and appropriateness of generated content, particularly crucial for low-resource languages where training data may be limited.

\paragraph{Educational Workflow:}
The integration of foundation models must complement existing language teaching methodologies. This includes developing interfaces that allow educators to monitor and guide the learning process, ensuring that automated adaptations align with pedagogical goals. Assessment mechanisms need to balance automated evaluation with human oversight, particularly for nuanced aspects of language acquisition like cultural competency~\cite{steele2000language}.

\paragraph{Resource Management:}
For low-resource language contexts, considerations must be made for environments with limited technological infrastructure. This includes developing offline functionality and optimizing model performance for resource-constrained settings. The system should be designed to gracefully handle intermittent connectivity while maintaining learning continuity.

\paragraph{Quality Assurance:}
Maintaining educational effectiveness requires robust quality assurance processes. This includes automated testing of generated content, collection of user feedback, and regular assessment of learning outcomes. For low-resource languages, additional verification may be needed to ensure cultural and linguistic accuracy, potentially involving community experts in the review process.

\paragraph{Community Engagement:}
Successful integration requires active participation from the language community. This includes establishing channels for native speakers and cultural experts to provide feedback and validation of educational content. Such engagement helps ensure that the adaptive learning system remains aligned with community needs and cultural values while supporting language preservation efforts.

\section{Conclusion and Future Challenges}
Incorporating large language models into low-resource language education poses significant educational challenges and opportunities. 
One major issue is the development of teaching methodologies to effectively utilize LLM capabilities, such as generating language exercises or providing feedback. Educators must develop pedagogical strategies that leverage LLM strengths while addressing their limitations, such as occasional inaccuracies or lack of context-specific nuances. 
Additionally, there is the challenge of integrating these technologies into existing educational frameworks in a way that complements traditional teaching methods rather than replacing them. This integration also requires substantial training for educators, who must become adept at using LLMs and understanding their potential biases and shortcomings. 
Furthermore, evaluating the effectiveness of LLMs in improving language competencies in diverse educational settings such as adaptive learning~\cite{wen2024ai} remains a critical challenge, necessitating ongoing research and adaptation to ensure that these tools are making a positive impact on learning outcomes.



\bibliography{anthology,custom}
\bibliographystyle{acl_natbib}




\end{document}